\documentclass[journal]{IEEEtran}
\ifCLASSINFOpdf
\else
\fi
\usepackage{graphicx}
\usepackage{amsmath}
\usepackage{cite}
\begin{document}
%

\title{Saliency Detection for Stereoscopic Images Based on Depth Confidence Analysis and Multiple Cues Fusion}
%
%
%

\author{Runmin Cong,~\IEEEmembership{Student Member,~IEEE,} Jianjun Lei,~\IEEEmembership{Member,~IEEE,} Changqing Zhang,\par
        Qingming Huang,~\IEEEmembership{Senior Member,~IEEE,}
        ~Xiaochun Cao,~\IEEEmembership{Senior Member,~IEEE,}
        and Chunping Hou
\thanks{Manuscript received January, 2016. This work was partially supported by the Natural Science Foundation of China (No. 61271324, 61520106002, 61471262, 91320201). Copyright (c) 2015 IEEE. Personal use of this material is permitted. However, permission to use this material for any other purposes must be obtained from the IEEE by sending a request to pubs-permissions@ieee.org.}
\thanks{R. Cong,  J. Lei (corresponding author), and C. Hou are with the School of Electronic Information Engineering, Tianjin University, Tianjin 300072, China (e-mail: rmcong@tju.edu.cn; jjlei@tju.edu.cn; hcp@tju.edu.cn).}
\thanks{C. Zhang is with the School of Computer Science and Technology, Tianjin University, Tianjin 300072, China (e-mail: zhangchangqing@tju.edu.cn).}
\thanks{Q. Huang is with Institute of Computing Technology, Chinese Academy of Sciences, Beijing 100190, China (e-mail: qmhuang@ucas.ac.cn).}
\thanks{X. Cao is with Institute of Information Engineering, Chinese Academy of Sciences, Beijing 100093, China (e-mail: caoxiaochun@iie.ac.cn).}}

%
%

\markboth{IEEE SIGNAL PROCESSING LETTERS,~Vol.~xx, No.~xx, xxxx~2016}%
{Shell \MakeLowercase{\textit{et al.}}: Bare Demo of IEEEtran.cls for IEEE Journals}
%



\maketitle

\begin{abstract}
Stereoscopic perception is an important part of human visual system that allows the brain to perceive depth. However, depth information has not been well explored in existing saliency detection models. In this letter, a novel saliency detection method for stereoscopic images is proposed. Firstly, we propose a measure to evaluate the reliability of depth map, and use it to reduce the influence of poor depth map on saliency detection. Then, the input image is represented as a graph, and the depth information is introduced into graph construction. After that, a new definition of compactness using color and depth cues is put forward to compute the compactness saliency map. In order to compensate the detection errors of compactness saliency when the salient regions have similar appearances with background, foreground saliency map is calculated based on depth-refined foreground seeds selection mechanism and multiple cues contrast. Finally, these two saliency maps are integrated into a final saliency map through weighted-sum method according to their importance. Experiments on two publicly available stereo datasets demonstrate that the proposed method performs better than other 10 state-of-the-art approaches.
\end{abstract}


%
\IEEEpeerreviewmaketitle

\section{Introduction}

\IEEEPARstart{A}{CCORDING} to the principle of human visual perception, people would like to place more attention on the region which stands out from the image background. Saliency detection is becoming an increasingly important branch in computer vision due to its various applications in object detection and recognition \cite{Ren2014}, \cite{Wang2014}, image retrieval \cite{Gao2015}, image compression \cite{Han2006},\cite{HYH}, and image retargeting \cite{Yoo2013}, \cite{Gallea2014}. Generally, saliency detection models can be categorized into data-driven bottom-up model and task-driven top-down model \cite{FANG1}. In this letter, we focus on bottom-up saliency detection models.\par

Saliency detection aims to effectively highlight salient regions and suppress background regions. By far, many saliency detection methods for RGB image have been put forward, which integrate different visual cues to compute the saliency map. As a pioneer, Itti \emph{et al.} \cite{Itti1998} presented a multi-scale saliency detection model, which computes center-surround differences using three visual features including color, intensity, and orientation. Cheng \emph{et al.} \cite{Cheng2011}, \cite{Cheng2015} proposed a region-based saliency model that measures the global contrast between the target regions with respect to all other regions in the image. In \cite{Yang2013}, graph-based manifold ranking was introduced into saliency detection model, which ranks the differences between salient object and background. Wei \emph{et al.} \cite{Wei2015} proposed a multiple saliency methods fusion framework based on Dempster-Shafer Theory (DST), and improved the performance of saliency detection. Sun \emph{et al.} \cite{Sun2015} cast saliency detection into a Markov problem, which integrate background prior and Markov absorption probability on a weighted graph. Recently, some new methods combine multiple cues into saliency detection model to achieve a better result. In \cite{Zhou2015}, Zhou \emph{et al.} found that compactness and local contrast are complementary to each other, and proposed a saliency detection method that integrates compactness and local contrast. In \cite{LiTIP2015}, Li \emph{et al.} proposed a novel label propagation method for saliency detection, which integrates backgroundness, objectness, and compactness cues.\par

Most previous works on saliency detection are focused on 2D images, which mainly concentrate on RGB color information while ignoring depth/disparity cue \cite{FHZ}, \cite{lei1}. In fact, 3D visual information can supply a useful cue for saliency detection \cite{fang2}, \cite{lei2}. Niu \emph{et al.} \cite{Niu2012} proposed a saliency model for stereoscopic images based on global disparity contrast and domain knowledge in stereoscopic photography. Desingh \emph{et al.} \cite{Desingh2013} computed the stereo saliency by fusing depth saliency with 2D saliency models, which through non-linear support vector regression. In \cite{Guo2014}, the disparity maps are used to refine the 2D saliency model and maintain the consistency between the stereo matching and saliency maps. Ju \emph{et al.} \cite{Ju2014}, \cite{Ju2015} proposed a depth-aware method for saliency detection, using an anisotropic center-surround difference (ACSD) measure. In addition, they construct a large dataset for stereo saliency detection, which includes 1985 stereo images and estimated depth maps.\par

Considering the effectivity of integrating depth information, we propose a novel saliency detection model for stereoscopic images in this letter. The main contributions can be summarized as follows: 1) A good depth map can be benefit for the saliency detection no matter how it produced. Consequently, it is vital that construct a measure to describe the quality of depth map. According to the observation of depth distribution, a confidence measure for depth map is proposed to reduce the influence of poor depth map on saliency detection; 2) A novel model for compactness integrating color and depth information is put forward to compute the compactness saliency; 3) A foreground seeds selection mechanism based on depth-refined is presented. The saliency is measured by contrast between the target regions with seed regions, which integrate color, depth, and texture cues.\par

\section{Proposed Method}

\begin{figure}[htbp]
\centering
\includegraphics[width=8.2cm,height=3.7cm]{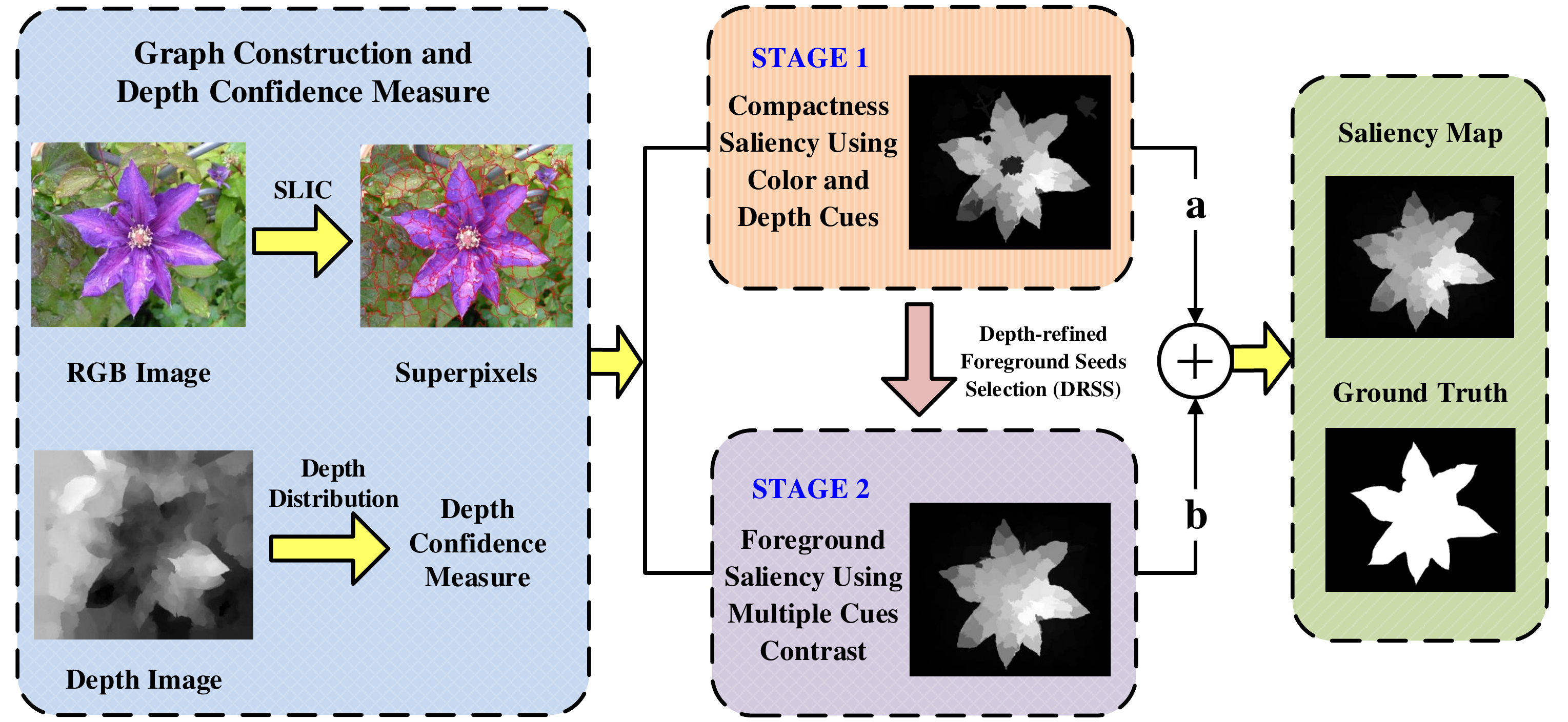}
\caption{Flowchart of the proposed method. }
\label{fig1}
\end{figure}

The flowchart of the proposed stereo saliency detection method is depicted as Fig. \ref{fig1}. There are four parts in our method. Firstly, a novel confidence measure is calculated to evaluate the reliability of depth map, and used in the following processes. The depth confidence measure can reduce the influence of poor depth map on saliency detection. Simultaneously, RGB image is abstracted into superpixels and represented as a graph. Then, compactness saliency based on color and depth cues is calculated. Further, foreground seeds are selected by combining compactness saliency result and depth map. Taking color, depth, and texture cues into consideration, a multiple cues contrast saliency detection method based on foreground seeds is proposed. At last, compactness and foreground saliency map are weighted to obtain the final saliency map.\par

\subsection{Depth Confidence Measure}

\begin{figure}[!t]
\centering
\includegraphics[width=1\linewidth]{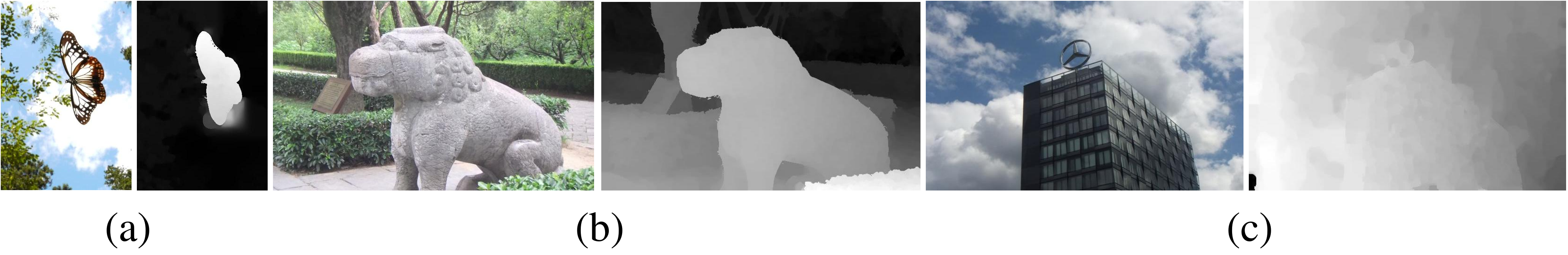}
\caption{Different quality of depth maps. (a) Good depth map, $\lambda_{d}=0.8014$. (b) Common depth map, $\lambda_{d}=0.3890$. (c) Poor depth map, $\lambda_{d}=0.0422$. }
\label{fig2}
\end{figure}

The quality of depth map is very important for the using of depth cue. A good depth map can provide accurate depth information which benefits for saliency detection. In contrast, the poor depth map may cause a wrong detection. Thus, a depth confidence measure is proposed in this letter to evaluate the reliability of depth map. We found that a good depth map often owns clear hierarchy, and the salient object can be highlighted from the background. Fig. \ref{fig2} shows some examples of different quality depth maps. \par
We rank the input depth map into three grades roughly, namely good, common, and bad. Based on the observation of depth statistical characteristic, we found that the values are usually concentrate on lower part for good depth map, whereas the distribution of poor depth map tends to concentrate on relatively larger values. Therefore, the mean value of the whole depth image is an effective parameter to tell them apart. In statistics, coefficient of variation is used to evaluate the dispersion degree of the data. It is observed that the poor depth map appears strong concentration compared with other cases. Thus, the coefficient of variation is introduced in our confidence measure. In addition, there is a more random distribution for a common depth map, therefore, the depth frequency entropy is defined to evaluate the randomness of an input depth image. According to these observations of depth distribution, we define the depth confidence measure as follows.
\par

\begin{equation}
\lambda_{d}=\exp((1-m_{d})\cdot\emph{CV}\cdot\emph{H})-1
\end{equation}
where $m_{d}$ is the mean value of the whole depth image, $\emph{CV}=m_{d}/\sigma_{d}$ is coefficient of variation, $\sigma_{d}$ is the standard deviation for depth image, and $\emph{H}$ is the depth frequency entropy, which denotes the randomness of depth distribution, and can be defined as follows.
\begin{equation}
\emph{H}=-\sum_{i=1}^LP_{i}\log(P_{i})
\end{equation}
where $P_{i}=n_{i}/n_{\Sigma}$, $n_{\Sigma}$ is the number of pixels in the depth map, $n_{i}$ is the number of pixels that belong to the region level $r_{i}$, and $L$ is the levels of depth map. In this letter, the input depth map is normalized to $[0,1]$ firstly. Then, $L-1$ thresholds, namely $T_{k}$, are used to divide the depth map into $L$ levels. A larger $\lambda_{d}$ corresponds to more reliable of the input depth map.
\subsection{Graph Construction}
The input RGB image is abstracted into homogenous and compact regions using SLIC superpixel segmentation method \cite{Achanta2012}. The number of superpixel $N$ is set to 200 in experiments. Then, we construct a graph $G=(V,E)$, where $V$ represents the set of nodes which corresponds to the superpixels generated by SLIC method, and $E$ denotes the set of links between adjacent nodes.\par
In this work, the Euclidean distance $l_{ij}$ in \emph{CIE Lab} color space and depth difference $d_{ij}$ between superpixels $v_{i}$ and $v_{j}$ are defined as
\begin{equation}
l_{ij}=\|\textbf{\emph{c}}_{i}-\textbf{\emph{c}}_{j}\|
\end{equation}
and
\begin{equation}
d_{ij}=|d_{i}-d_{j}|
\end{equation}
where $\textbf{\emph{c}}_{i}$ is the mean color value of superpixel $v_{i}$, and $d_{i}$ denotes the mean depth value of superpixel $v_{i}$. The similarity between two superpixels $a_{ij}$ is defined as
\begin{equation}
a_{ij}=\exp(-\frac{l_{ij}+\lambda_{d}\cdot{d_{ij}}}{\sigma^{2}})
\end{equation}
where $\sigma^{2}$ is a parameter to control strength of the similarity, and it is set to 0.1 in all our experiments. The affinity matrix $\textbf{\emph{W}}=[w_{ij}]_{N\times N}$ is defined as the similarity between two adjacent superpixels\par
\begin{equation}
w_{ij}=
\begin{cases}
a_{ij} & \text{if $j\in\Omega_{i}$}\\
0 & \text{otherwise}
\end{cases}
\end{equation}
where $\Omega_{i}$ is the set of neighbors of superpixel $v_{i}$.
\subsection{Compactness Saliency Using Color and Depth Cues}
Intuitively, salient regions have compact spread in spatial domain, whereas the colors of background regions have a larger spread over the whole image \cite{Zhou2015}. In addition, the values of depth exhibit limited compactness, that is, the depth values of salient regions are more likely to have a centralized distribution near the center of image. Motivated by this, we calculate the compactness saliency using color and depth cues. To obtain more precise result, we first propagate the similarity $a_{ij}$ using manifold ranking method \cite{Zhou2004}.\par
Zhou \emph{et al.} proposed a compactness metric using color spatial information in \cite{Zhou2015}. In this letter, a novel measure is proposed to calculate the color and depth-based compactness. The saliency map based on compactness is defined as
\begin{equation}
S_{CS}(i)=[1-norm(cc(i)+dc(i))]\cdot{Obj(i)}
\end{equation}
where $cc(i)$ is the color-based compactness of superpixel $v_{i}$, $dc(i)$ is the depth-based compactness of superpixel $v_{i}$, and $norm(x)$ is a function that normalizes $x$ to $[0,1]$ using min-max normalization. Considering the importance of location for saliency detection, objectness measure \cite{Alexe2012} $Obj(i)$ is introduced in our model to evaluate the probability of superpixel $v_{i}$ that belongs to an object. The color and depth-based compactness are defined as
\begin{equation}
cc(i)=\frac{\sum_{j=1}^\emph{N} a_{ij}\cdot{n_{j}\cdot{\|\textbf{\emph{b}}_{j}-\boldsymbol{\mu}_{i}\|}}}{\sum_{j=1}^\emph{N} a_{ij}\cdot n_{j}}
\end{equation}
and
\begin{equation}
dc(i)=\frac{\sum_{j=1}^\emph{N} a_{ij}\cdot{n_{j}\cdot{\|\textbf{\emph{b}}_{j}-\textbf{\emph{p}}\|}\cdot \exp(-\frac{\lambda_{d}\cdot{d_{i}}}{\sigma^{2}})}}{\sum_{j=1}^\emph{N} a_{ij}\cdot n_{j}}
\end{equation}
where $n_{j}$ denotes the number of pixels that belong to superpixel $v_{j}$, which emphasizes the impact of larger region, $\textbf{\emph{b}}_{j}=[b^{x}_{j},b^{y}_{j}]$ is the centroid coordinate of superpixel $v_{j}$, $\textbf{\emph{p}}=[c_{x},c_{y}]$ is the spatial position of the image center, and the spatial mean $\boldsymbol{\mu}_{i}=[\mu^{x}_{i},\mu^{y}_{i}]$ is defined as
\begin{equation}
\mu^{x}_{i}=\frac{\sum_{j=1}^\emph{N} a_{ij}\cdot{n_{j}}\cdot b^{x}_{j}}{\sum_{j=1}^\emph{N} a_{ij}\cdot n_{j}}
\end{equation}
and
\begin{equation}
\mu^{y}_{i}=\frac{\sum_{j=1}^\emph{N} a_{ij}\cdot{n_{j}}\cdot b^{y}_{j}}{\sum_{j=1}^\emph{N} a_{ij}\cdot n_{j}}
\end{equation}

\subsection{Foreground Saliency Using Multiple Cues Contrast}
Although compactness saliency detection method is active on some level, there are some limitations. When the salient regions have similar appearances with background, the regions may be wrongly detected. Hence, a foreground saliency detection method based on multiple cues contrast is proposed to mitigate this problem.\par
\begin{figure}[htbp]
\centering
\includegraphics[width=6.4cm,height=1.9cm]{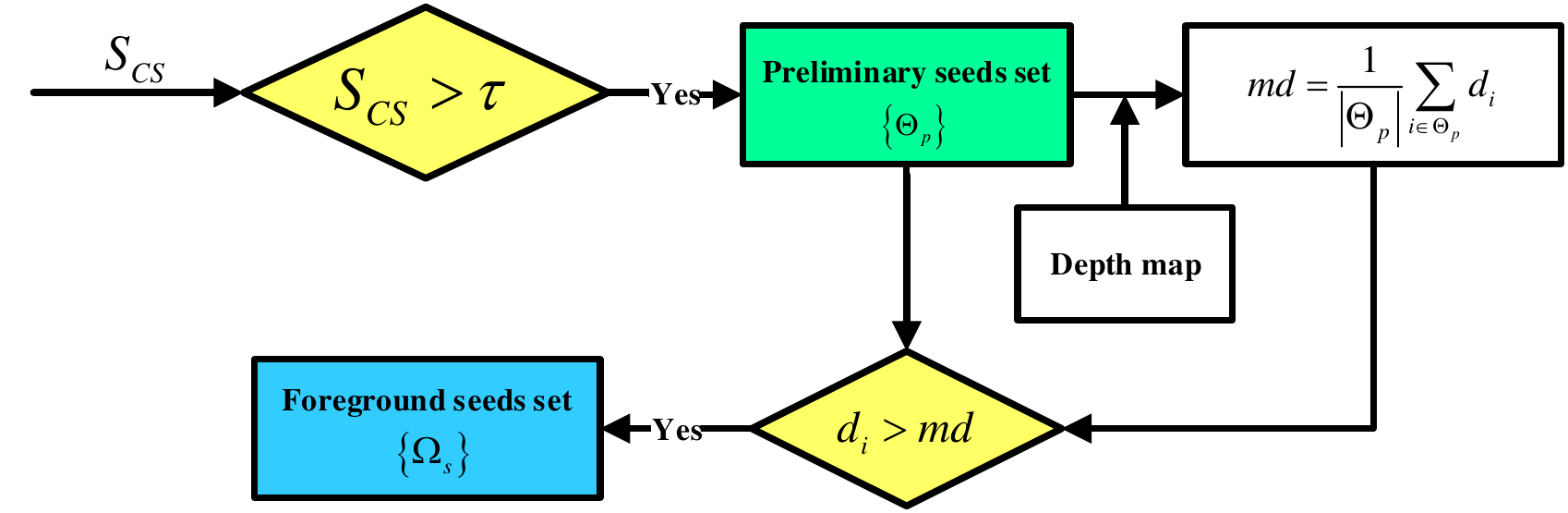}
\caption{Flowchart of depth-refined foreground seeds selection mechanism. }
\label{fig3}
\end{figure}

Traditionally, foreground seeds are selected just based on the preliminary saliency map. Considering the importance of depth information, a novel selection mechanism for foreground seeds is introduced, in which the seeds should own larger values of compactness saliency and depth simultaneously. Therefore, a depth-refined foreground seeds selection method (DRSS) is proposed, and the flowchart is shown in Fig. \ref{fig3}. Firstly, preliminary seeds are determined by thresholding segmentation using compactness saliency map, where the threshold $\tau$ is set to 0.5. Then, the mean depth value of preliminary seeds is used to refine the preliminary seeds, and obtain the final foreground seeds set.\par
Next, we calculate the contrast of each superpixel with the foreground seeds based on multiple cues, which include color, depth, texture, and position information. A superpixel is more likely to be salient if it is more similar to the foreground seeds. The foreground saliency is computed as follows.
\begin{equation}
S_{fg}(i)=\sum_{j\in\Omega_{s}}[a_{ij}\cdot D_{t}(i,j)\cdot \exp(-\|\textbf{\emph{b}}_{i}-\textbf{\emph{b}}_{j}\|/\sigma^{2})\cdot n_{j}]
\end{equation}
where $\Omega_{s}$ is the set of foreground seeds, $\|\textbf{\emph{b}}_{i}-\textbf{\emph{b}}_{j}\|$ denotes the Euclidean distance between position of superpixels, and $D_{t}(i,j)$ is the texture similarity between superpixels using LBP feature \cite{Ojala2002}, which defined as follows.
\begin{equation}
D_{t}(i,j)=\frac{|\textbf{\emph{k}}^{T}_{i}\textbf{\emph{k}}_{j}|}{\|\textbf{\emph{k}}_{i}\|\|\textbf{\emph{k}}_{j}\|}
\end{equation}
where $\textbf{\emph{k}}_{i}$ is LBP histogram frequency of superpixel $v_{i}$. To avoid the problem that saliency map highlights object boundaries rather than the entire region, manifold ranking method is used to propagate the foreground saliency map. At last, the map after propagation is normalized to  $[0,1]$, and the final foreground saliency map $S_{FS}$ is obtained.
\subsection{Saliency Map Integration}
The compactness and foreground saliency maps are complementary to each other. Considering the foreground saliency map is based on the compactness saliency result, we integrate these two saliency map into a final saliency map through weighted-sum method.
\begin{equation}
S=\gamma\cdot S_{CS}+(1-\gamma)\cdot S_{FS}
\end{equation}
where $\gamma$ balances the compactness saliency map $S_{CS}$ and foreground saliency map $S_{FS}$.
\section{Experimental Results}
We evaluate the performance of our method on two datasets: NJU-400 \cite{Ju2014} and NJU-1985 \cite{Ju2015}, which include RGB images, depth maps, and pixel-wise ground truth annotations. The performance is evaluated using precision-recall curve, F-measure, and Mean Absolute Error (MAE) \cite{Cheng2011,FYM3,Qin2015}. The precision-recall curve is obtained by binarizing  the saliency map using thresholds in the range of 0 and 255. In all experiments, we set the parameters $L=3$, $T_{1}=0.4$, $T_{2}=0.6$, and $\gamma=0.8$.\par

\subsection{Performance Comparison}
\begin{figure}[htbp]
\centering
\includegraphics[width=8cm,height=6.5cm]{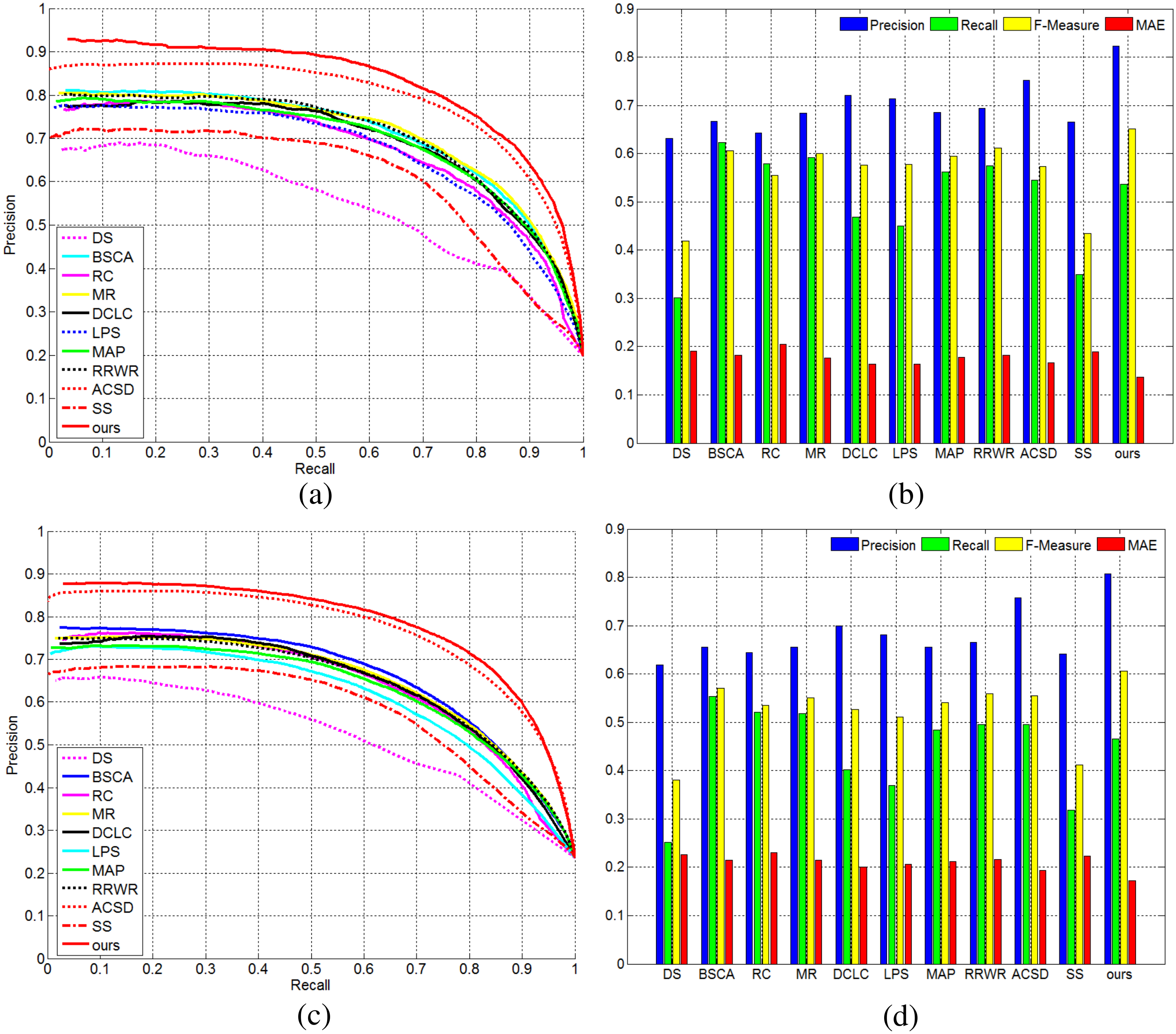}
\caption{The quantitative comparison of proposed method with 10 state-of-the-art methods. (a) Precision-recall curves of different methods on NJU-400 dataset. (b) Average precision, recall, F-measure, and MAE of different methods on NJU-400 dataset. (c) Precision-recall curves of different methods on NJU-1985 dataset. (d) Average precision, recall, F-measure, and MAE of different methods on NJU-1985 dataset.}
\label{fig4}
\end{figure}

\begin{figure}[htbp]
\centering
\includegraphics[width=8cm,height=3.8cm]{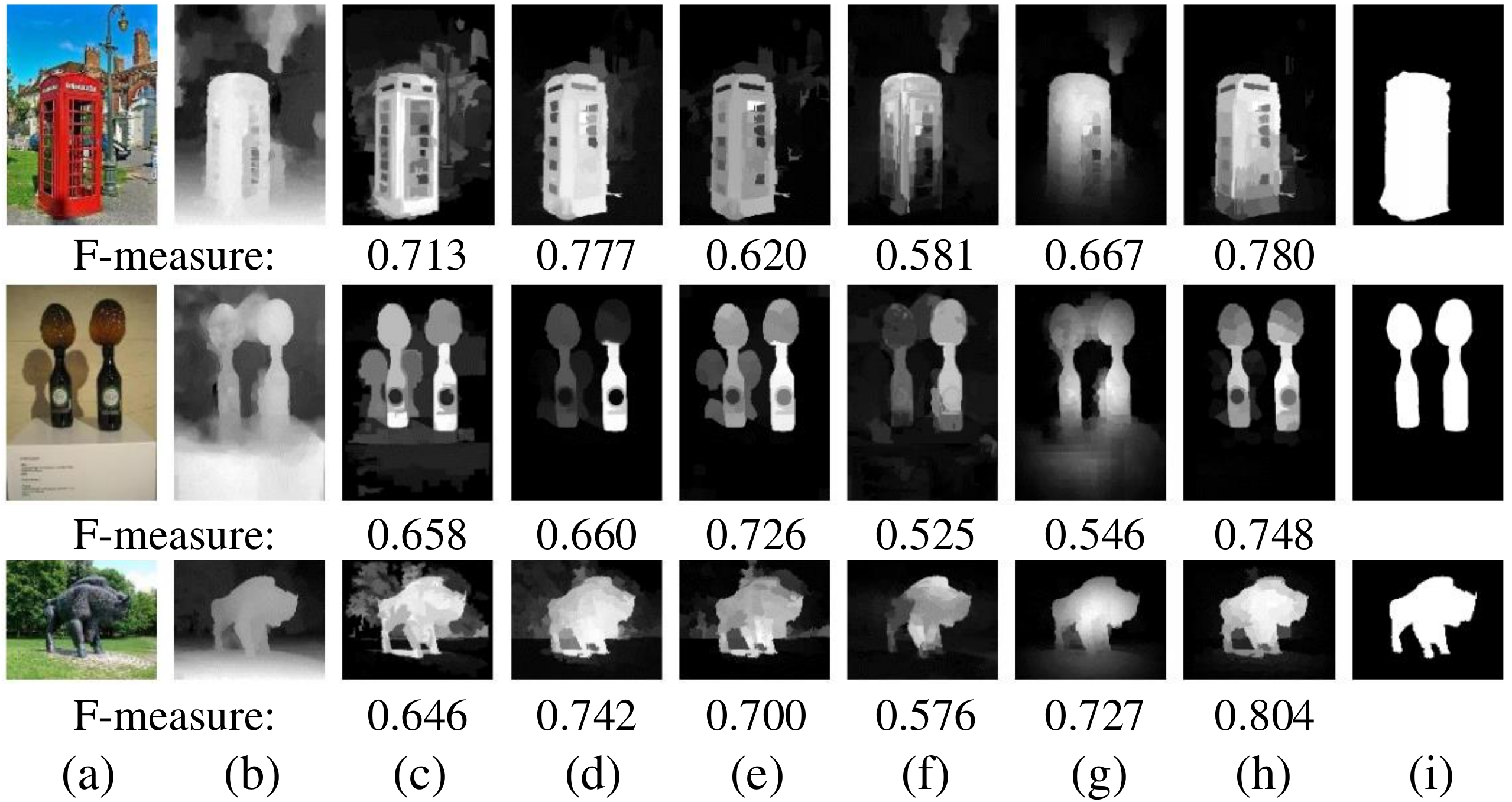}
\caption{Visual comparison of saliency maps. (a) Input RGB image. (b) Input depth image. (c) RC \cite{Cheng2015}. (d) RRWR \cite{Li2015}. (e) DCLC \cite{Zhou2015}. (f) SS \cite{Niu2012}. (g) ACSD \cite{Ju2015}. (h) Ours. (i) Ground truth. }
\label{fig5}
\end{figure}

We compare our method with 8 state-of-the-art 2D methods: RC \cite{Cheng2015}, MR \cite{Yang2013}, DS \cite{Wei2015}, MAP \cite{Sun2015}, DCLC \cite{Zhou2015}, LPS \cite{LiTIP2015}, BSCA \cite{Qin2015}, RRWR \cite{Li2015}, and 2 stereo saliency detection methods: SS \cite{Niu2012}, ACSD \cite{Ju2015}. Fig. \ref{fig4} shows the evaluation results of the proposed method with 10 state-of-the-art methods on two datasets. On both datasets, the precision-recall curves show that the proposed method performed better than other methods. Similarly, the proposed method achieves the best performance in terms of the average precision, F-measure, and MAE compared with other approaches due to the depth confidence analysis and two-stage saliency computation mechanism. Taking the F-measure on the NJU-1985 dataset as an example, the F-measure values of two stereo saliency detection methods (ACSD and our method) are 0.5552 and 0.6055, respectively. Nevertheless, our method achieved a lightly lower average recall value than ACSD method. In the future work, the recall value should be improved on the premise of that the value of precision has been maintained. Fig. \ref{fig5} presents visual comparison of different saliency detection methods. The proposed method has more similar appearances with ground truth, and owns clear contour and uniform salient regions. Furthermore, the F-measure values in Fig. \ref{fig5} also demonstrate the effectiveness of our proposed method.\par
\subsection{Parameter Analysis}
\begin{figure}[htbp]
\centering
\includegraphics[width=7cm,height=3cm]{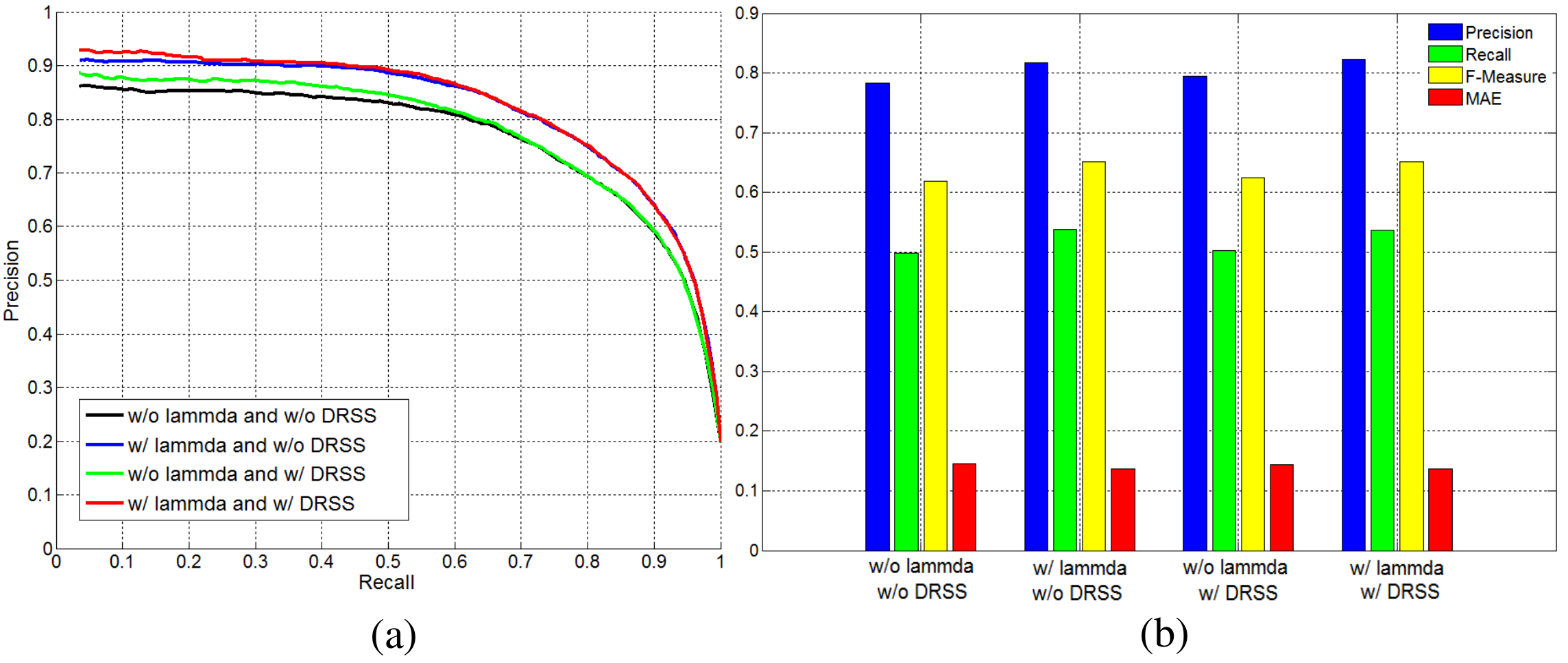}
\caption{Evaluation of different factors for final saliency detection on NJU-400 dataset. (a) Precision-recall curves of different factors. (b) Average precision, recall, F-measure, and MAE of different factors. }
\label{fig6}
\end{figure}

In this section, we evaluate the performance of our method under different factors. The precision-recall curves and quantitative indexes are shown in Fig. \ref{fig6}. In order to reduce the influence of poor depth map in stereo saliency detection, depth confidence measure $\lambda_{d}$ is introduced in our letter. Comparing the black line with the blue line in Fig. \ref{fig6}(a), it demonstrated that the performance with $\lambda_{d}$ is superior to not using depth confidence measure, and the same conclusion can be drawn from the comparison of the first two columns in Fig. \ref{fig6}(b). At the stage of foreground saliency detection, a novel foreground seeds selection mechanism using depth information, namely DRSS, is proposed to acquire more accurate foreground seeds. As shown in Fig. \ref{fig6}, the DRSS process can achieve higher precision rate, and improve the performance of the final result according to the precision-recall curve.\par

\section{Conclusion}
In this letter, a novel saliency detection model for stereoscopic images based on depth confidence analysis and multiple cues fusion is presented. First, the quality of depth map is considered when introduces the depth information into the saliency model, and a depth confidence measure is proposed to evaluate the reliability of depth map. In addition, a novel model for compactness integrating color and depth information is proposed to compute the compactness saliency. To achieve more robust saliency detection result, a foreground saliency detection method based on multiple cues contrast is proposed, which includes a novel depth-refined foreground seeds selection method. At last, weighted-sum method is used to generate the final saliency map. Experimental evaluation on two public benchmarks has validated the advantages of our approach.\par


%

\ifCLASSOPTIONcaptionsoff
  \newpage
\fi

\end{document}